\documentclass[11pt]{article}
\usepackage{acl2015}
\usepackage{times}
\usepackage{url}
\usepackage{latexsym}
\usepackage{amsthm}
\usepackage{graphicx}
\usepackage{hyperref}

\newcounter{examplecounter}
\newenvironment{example}{\begin{quote}%
    \refstepcounter{examplecounter}%
  (\arabic{examplecounter})%
  \quad
}{%
\end{quote}%
}

\newcommand*\samethanks[1][\value{footnote}]{\footnotemark[#1]}

\title{Extracting Biological Pathway Models From NLP Event Representations}

\author{Michael Spranger \thanks{These two authors contributed equally to this paper and the software system.} \\
	Sony Computer Science\\ 
	Laboratories Inc.\\
	Tokyo, Japan\\
  {\tt michael.spranger}\\{\tt @gmail.com} \\\And
  Sucheendra K. Palaniappan \samethanks \\
 INRIA, \\ Campus de Beaulieu,\\ 
 Rennes, France \\
  {\tt sucheendra.palaniappan}\\{\tt @inria.fr} \\\And
  Samik Ghosh \\
	 The Systems Biology Institute, \\
	 Minato-ku, \\
	 Tokyo, Japan\\
  {\tt ghosh@sbi.jp} \\}

\begin{document}
\maketitle

\begin{abstract}
This paper describes an an open-source software system for the automatic conversion of NLP event representations
to system biology structured data interchange formats such as SBML and BioPAX.
It is part of a larger effort to make results of the NLP community available for system biology pathway modelers. 
\end{abstract}
 
\section{Introduction}
Biological pathways represent important insights into the flow of information within a cell by encoding the sequence of interactions among various biological players (such as genes, proteins etc.) in response to certain stimuli (or spontaneous at times) which leads to a change in the state of the cell. Studying and analyzing these pathways is crucial to understanding biological systems.
 
Traditionally, pathways are represented as maps which are constructed and curated by expert curators who manually read numerous biomedical documents, comprehend and assimilate the knowledge into maps. This process is aided by a variety of graphical tools such as CellDesigner \cite{funahashi2008celldesigner}.

Such manual pathway curation comes with a number of problems. Most importantly: 1) the amount of time and therefore cost for detailed pathway maps is high. 2) As new research findings are published these pathway need to be updated or augmented.  Often, the speed at which molecular research is progressing, means it is hard to keep pathways in sync. 3) Many times the interpretation of details is left to the judgment of the curator, which leads to considerable variability of pathways.

Considering these limitations, there has been an increased emphasis 
on using Natural Language Processing (NLP) techniques for automated pathway curation. The BioNLP Shared Task - Pathway Curation (BioNLPST-PC) competition \cite{sharedtask2013,ohta2013overview} 
was focused on this specific problem. From the NLP perspective the extraction of biological knowledge
is posed as an event detection problem with standard NLP event detection algorithms used to extract the biological information 
from text \cite{Ananiadou2010}. 

Although there has been a lot of work on the problem of automatic pathway extraction from text, to our knowledge there has been little effort to make the extracted information available in standard pathway formats. The majority of pathway data is represented, stored and exchanged using standard formats such as SBML  \cite{hucka2003systems} and BioPAX \cite{demir2010biopax}. Contrary to these formats existing NLP extraction systems often use a data format called the ``standoff format", to represent their results. While the standoff format is often described as easily convertible into SBML and BioPAX, no actual software seems to exist to automate this conversion. This paper tries to fill this gap by describing a software system for the conversion of NLP event representations to the system biology structured data interchange formats SBML and BioPAX. We also provide open sourced software tools {\tt st2sbml} and {\tt st2biopax} to convert from stand-off to SBML/BioPAX format. The software tools and additional information about the contents of this paper can be found on our supplementary webpage\footnote{\url{https://github.com/sbnlp/standoff-conversion}}.

\section{NLP Event Representations}
Existing NLP systems often use an event representation format comprised of a 
set of annotation rules and file formats to represent pathway events and entities \cite{kim2011overview}.
For the purpose of this paper we base ourselves on the standoff representation (ST) proposed 
for the BioNLP Shared Task 2011, 2013 \cite{sharedtask2013}. 

Annotations in ST link spans of texts through 
character offsets to {\tt entities} (e.g. Proteins, Genes etc.) and {\tt events} (Positive 
Regulation etc.). Events and entities are represented line by line with links between them. 

The following is an example sentence and a possible event representations.

\begin{example}
\label{e:example-1}
YAP modulates the phosphorylation of Akt1.
\end{example}

\begin{figure}
\includegraphics[width=\columnwidth]{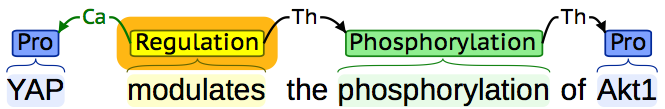}
\caption{Graphical representation of the event representation of Example \ref{e:example-1}}
\label{f:event}
\end{figure}

\begin{small}
\footnotesize
\begin{verbatim}
T1 Protein 0 3 YAP
T2 Protein 37 41 Akt1
T3 Regulation 4 13	modulates
T4 Phosphorylation 17 32	phosphorylation
E1 Phosphorylation:T4	Theme:T2
E2 Regulation:T3 Theme:E1 Cause:T1
\end{verbatim}
\end{small}

Each annotation starts with a unique annotation-ID. The annotations-IDs encodes the annotation type
in the first letter (T - text bound annotation, E - event annotation). This is followed by the annotation-type.
For instance, the text bound annotation T1 is of type protein, whereas T3 is of type Regulation.
Text bound annotations also encode the start and end position as well as the text they annotate. Text bound
annotation T1 for instance ranges from character 0 to character 3 of the annotated text and the actual text is
``YAP''.

Event annotations build on top of text bound annotation. The annotations-ID for an event is followed by an event-type 
and the reference to the text bound annotation. For instance, E1 is a Phosphorylation  
event and the corresponding text is T4 ``phosphorylation''. Additionally, event annotations encode roles. T2 is the 
theme of E1, which in this case means that ``Akt1'' is undergoing a phosphorylation. Events can also be used 
as theme. For example the theme of E2 is E1, which means that the phosphorylation is 
regulated by ``YAP''. Different roles are possible depending on the type of the event. 

\section{From Event Representations to SBML}

Systems Biology Markup Language, or short SBML \cite{hucka2003systems}, is a XML-based 
markup language to describe, store and communicate biological models. 
It is among the most widely used formats with numerous software support. SBML 
essentially encodes models using biological players called {\tt sbml:species}\footnote{We will refer to SBML vocabulary using the prefix ``sbml".}.
{\tt sbml:species} can participate in interactions, called {\tt sbml:reaction}. 
Species participate in interaction as {\tt sbml:reactant}, {\tt sbml:product} and {\tt sbml:modifier}.
The basic idea being that some quantity of reactant is consumed to produce a product.
Reactions are influenced by modifiers. 

SBML supports mathematical representations of the underlying dynamics of the reactions 
and is essentially used to simulate models. Due to this, there is no SBML vocabulary 
to specify different types of reactions (such as transcription, phosphorylation etc.) or 
species (such as protein, DNA etc.). Alternatively, species and reactions can be annotated 
and uniquely specified using MIRIAM resources and annotations \cite{novere2005minimum}.
We use controlled vocabulary from the Systems Biology Ontology (SBO) and 
the Gene Ontology (GO). This information is also useful to convert SBML 
files to other formats such as SBGN \cite{SBGN} using tools such as VANTED \cite{vanted}.

\begin{figure}
\begin{centering}
\includegraphics[width=.8\columnwidth]{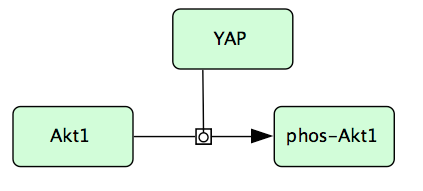}
\caption{Example \ref{e:example-1} converted into SBML (viewed with CellDesigner)}
\label{f:sbml}
\end{centering}
\end{figure}

Figure \ref{f:sbml} shows Example \ref{e:example-1} converted into an SBML model
using the mapping algorithm described in the following paragraphs.

\subsection{Mapping Algorithm}
The conversion of standoff formatted information to an SBML model consists 
of five steps.

\paragraph{Step 1: Initialize the Model}
Firstly, read the event annotation files and create a memory internal
representation of triggers and events. We initialize an empty SBML 
model with a single {\tt sbml:compartment} named ``default''.

\begin{table}	[t]
\begin{footnotesize}
\begin{tabular}{ p{1.5cm} p{2cm}  p{3.1cm} }
{\bf Standoff Entity} & {\bf SBO term} & {\bf SBO name}\\
\hline
Complex & SBO:0000253 & non-covalent complex\\
Gene\_or\_\\gene\_product & SBO:0000245 & macromolecule\\
Dna & SBO:0000251 & deoxyribonucleic acid\\
DnaRegion & SBO:0000251 & deoxyribonucleic acid\\
Drug &	SBO:0000247 & simple chemical\\
Ion & SBO:0000327 & non-macromolecular ion\\
Protein & SBO:0000252 & polypeptide chain\\
Rna & SBO:0000250 & ribonucleic acid\\
RnaRegion & SBO:0000250 & ribonucleic acid\\
Gene & SBO:0000354 & informational molecule segment\\
Small\\Molecule& SBO:0000247 & simple chemical\\
Simple\_\\molecule & SBO:0000247 & simple chemical\\
\end{tabular}
\caption{Mapping of Annotation-type to SBO term.}
\end{footnotesize}
\label{t:mapping-entity-sbo}
\end{table}

\paragraph{Step 2: Create {\tt\bf sbml:species}}
For each entity in the standoff format, a {\tt sbml:species} is added to 
the SBML model. This only applies to standoff entities that can be mapped
to an SBO term. Then the following is done 1) map the annotation-ID of the trigger to the id in 
the {\tt sbml:species}, 2) create a meta id by appending  "metaid\_0000" and annotation-ID;  meta id facilitates that annotations to this species can uniquely refer to it 3)
add the annotation-text as the name of the {\tt sbml:species}, 4) map the annotation-type to an 
SBO term and add to the {\tt sbml:species} (see Table 1)

For instance, the standoff line
\begin{small}
\begin{verbatim}
T2 Protein 37 41 Akt1
\end{verbatim}
\end{small}

will be mapped to 
\begin{small}
\begin{verbatim}
<species sboTerm="SBO:0000252"
 id="T2" name="Akt1"
 metaid="metaid_0000T2"
 compartment="default"/>
\end{verbatim}
\end{small}

On the other hand, a line such as 
\begin{small}
\begin{verbatim}
T39	 Entity 641 648	nucleus
\end{verbatim}
\end{small}
will not be used to 
create a species in the SBML model, because  ``Entity'' cannot be mapped 
to an SBO term. Here, ``nucleus" actually refers to a compartment which is
not directly deducible from the entity definition in the standoff format. To deal with
such cases, we need to take into account their role in Events something that is described 
in the next few paragraphs.

\begin{table}[t]
\label{t:mapping-event-sbo}
\begin{footnotesize}
\begin{tabular}{ p{2.3cm} p{2cm}  p{2.5cm}}
{\bf Standoff Event} & {\bf SBO/GO term} & {\bf SBO/GO name}\\
\hline
Conversion & SBO:0000182 & conversion \\
\hspace{0.1cm}Acetylation & SBO:0000215 & acetylation\\
\hspace{0.1cm}Deacetylation & GO:0006476 & Protein Deacetylation\\
\hspace{0.1cm}Methylation & SBO:0000214 & Methylation\\
\hspace{0.1cm}Demethylation & GO:0006482 & Protein Demethylation\\
\hspace{0.1cm}Phosphorylation & SBO:0000216 & phosphorylation\\
\hspace{0.1cm}Dephosphorylation & SBO:0000330 & Methylation\\
\hspace{0.1cm}Ubiquitination & SBO:0000224 & Ubiquitination\\
\hspace{0.1cm}Deubiquitination & GO:0016579 & Protein Deubiquitination\\
Degradation & SBO:0000179 & degradation \\
Catabolism & GO:0009056 & Catabolic Process \\
Catalysis & SBO:0000172 & Catalysis\\
Protein\_\\catabolism & GO:0009056 & Catabolic Process\\
Association & SBO:0000177 & non-covalent binding \\
Binding & SBO:0000177 & non-covalent binding\\
Dissociation & SBO:0000180 & dissociation\\
Regulation & GO:0065007 & biological regulation \\
\hspace{0.1cm}Positive\_ & GO:0048518 & positive regulation\\
\hspace{0.1cm}regulation\\
\hspace{0.2cm}Activation & SBO:0000412 & biological activity \\
\hspace{0.1cm}Negative\_ & GO:0048519 & negative regulation \\
\hspace{0.1cm}regulation\\
\hspace{0.2cm}Inactivation &  SBO:0000412 & biological activity\\
Gene\_\\expression & GO:0010467 & Genetic Production\\
\hspace{0.1cm}Transcription & SBO:0000183 & Transcription\\
\hspace{0.1cm}Translation & SBO:0000184 & Translation\\
Localization & GO:0051179 & Localization\\
\hspace{0.2cm}Transport & SBO:0000185 & Transport Reaction\\
Pathway & SBO:0000375 & Process
\end{tabular}
\caption{Mapping of annotation-type to SBO/GO term.}
\end{footnotesize}
\end{table}

\paragraph{Step 3: Create {\tt\bf sbml:reaction}}

Most events are added to the SBML model as {\tt sbml:reaction}.
For instance, the text trigger and event annotation corresponding to 
E1 in Example \ref{e:example-1} result in the following SBML description

\begin{small}
\begin{verbatim}
<reaction metaid="metaid_0000E1" 
          sboTerm="SBO:0000216" 
          id="E1" 
          name="Phosphorylation" 
          reversible="false">
 <annotation> ... </annotation>
</reaction>
\end{verbatim}
\end{small}

The SBO/GO term is assigned according to the mapping depicted in Table 2. 
The reaction id is based on the event id (E10). The metaid of the form "metaid\_0000 + id" is also added
and the {\tt sbml:reaction} name is the event-type. Lastly, all reactions are constructed as non reversible.

In a second step {\tt sbml:reactant}, {\tt sbml:product} and {\tt sbml:modifier} are added to 
SBML reactions based on the roles of events.

\begin{description}
\item[{\tt Theme}] is the entity that undergoes the effects of the event. It is mapped to the 
{\tt sbml:reactant} of the SBML reaction. For this a reactant reference is created and the 
species corresponding to the entity is linked to that reference via the id of the species (annotation-id of the entity).
\item[{\tt Product}] can be specified for Binding, Dissociation\footnote{In data used for evaluation we also encountered Dissociation events with {\tt Participant} and {\tt Complex} roles. They are mapped to {\tt sbml:product} and {\tt sbml:reactant} respectively.} and Conversion events. Product is mapped to
{\tt sbml:product} of the corresponding reaction. The entities appearing in the product role are used 
for creating a product reference with the same entity.
\item[{\tt Cause}] is an entity/event causing the event. Cause is eventually mapped to entities which are then mapped to the reaction
as {\tt sbml:modifier} (via modifier reference).

\end{description}
Information in {\tt Site} (which describes the site on the Theme entity that is modified in
the event) is added to the ``Notes" section of the SBML reaction as there seems to be no direct way to represent this information in SBML. Notes are human-readable annotations that can be added to SBML reactions.

\paragraph{Step 4: Handle Localization and Transport Events}
Localization and Transport events are handled differently from other events. They occur with additional roles besides {\tt Theme}.

\begin{description}
\item[ {\tt AtLoc}] describes the location/compartment at which the entity/species is located not an actual reaction. Hence, 
localization events with {\tt AtLoc} roles do not end up as reactions in SBML. Instead, first we check if a {\tt sbml:compartment} 
described by the {\tt AtLoc} role exists, else a new {\tt sbml:compartment} is created (see the nucleus example discussed 
earlier). Next, the compartment of the theme entity of the event is set to the corresponding {\tt sbml:compartment}.

\item[{\tt FromLoc/ToLoc}] Transport and Localization events can also include {\tt FromLoc} and {\tt ToLoc} roles which
describes the transport of the theme entity/species from some location/compartment to another. Consequently, 
we create a reaction where the {\tt Theme} entity/species starts out in the compartment described by {\tt FromLoc} ({\tt sbml:reactant}) and 
ends up in the compartment described by the {\tt ToLoc} ({\tt sbml:product}) role.
If the {\tt FromLoc/ToLoc} {\tt sbml:compartment} does not exist when creating 
the {\tt sbml:reaction}, a new {\tt sbml:compartment} is created corresponding to {\tt FromLoc/ToLoc}.
\end{description}

\paragraph{Step 5: Handle Gene Expression Events}
We model Gene expression events (e.g. Transcription and Translation) as reactions in SBML. 
However, this class of reactions does not have the {\tt sbml:reactant} role.
For Transcription events (process in which a gene 
sequence is copied to produce RNA) if the type of {\tt Theme} is RNA, it gets mapped to 
{\tt sbml:product}. If the type of {\tt Theme} is DNA, then it gets mapped to the 
{\tt sbml:modifier} of the Transcription {\tt sbml:reaction}.  

Translation events are handled in a similar manner.
 
\paragraph{Step 6: Handle Regulation Events}
In principle regulation events such as Positive/Negative Regulation, Activation
and Inactivation can be handled as described in Step 3 when the {\tt Theme} and
{\tt Cause} are species. If {\tt Theme} and {\tt Cause} are species then they are added to
a regulation reaction as reactant and modifier respectively. 

However, the standoff format definition also allows regulation events where 
{\tt Theme} and {\tt Cause} are themselves events\footnote{In some of the data
used to test our conversion we also encountered Catalysis events which had event themes. They are handled
exactly as Positive Regulation events.}. For example, the following 
standoff lines describe a Positive regulation of a Phosphorylation event. 

\begin{small}
\begin{verbatim}
T14	Protein 776 782	eIF-4E
T15	Protein 852 859	insulin
T43	Phosphorylation 820 835	phosphorylation
T44	Positive_regulation 839 848	
    increased
E21	Phosphorylation:T43 Theme:T14
E22	Positive_regulation:T44 Cause:T15 
    Theme:E21
\end{verbatim}
\end{small}

If the {\tt Theme} is an event, then we do not create a reaction but simply add
the {\tt Cause} entity as a modifier to the reaction corresponding to the {\tt Theme} event of the 
regulation. For the example above this means that the Phosphorylation reaction E21
is positively regulated (modified) by insulin (T15).

In reality though things are a bit more complicated since the {\tt Theme} event
might itself not exist as a reaction. For instance, there could be an event description as follows:

\begin{small}
\begin{verbatim}
E23    Positive_regulation:T35 Cause:T21 
		Theme:E13
E13    Positive_regulation:T36 Theme:E21
\end{verbatim}
\end{small}

Here, the event E23 has Theme E13, which itself is a Positive regulation with Theme E21. However, 
E13 itself does not correspond to a reaction. In this case the algorithm recursively tracks down 
the {\tt Theme} event across multiple event annotations until it finds an event that exists in the 
SBML model as a reaction (In this case E21 is identified as the Theme for E23). 

In case the {\tt Cause} is an event, the product of the {\tt Cause} event is used as a modifier. 
If the reaction corresponding to the Cause event does not have a product yet, then a
corresponding product species is first created and added to the model.

\paragraph{Step 7: Optional Cleanup and Annotation Operations}
As a last step optional cleanup/enhancement operations can be performed. 
They can be used to ensure consistency of the
resulting SBML model.

\begin{description}
\item[Add UniProt information] We use the annotation\\-text to retrieve information
about species from UniProt. The UniProt ID is added as controlled vocabulary to
the corresponding SBML species. Other information is added as XML annotation
and XHTML notes. This includes information about alternate names, gene names,
gene ids where available and appropriate.
\item[Remove unused species] Not all entities end up as products, reactants or modifiers
of an SBML reaction. In many cases, the named entity recognizer might recognize some
entity but no links to events is established. However, the entities might have been
added to the model (see Step 1). Entities not partaking in any reaction can be removed 
automatically.

\item[Complete reactions] 
Our software supports automatic adding of products and reactants for reactions that 
were not explicitly annotated in that way. For instance, all phosphorylation events
can extended with corresponding {\tt sbml:product} species. The completion takes
into account that certain reactions such as Gene expression reactions do not have
reactants.

Here is an example of what we mean. For a Phosphorylation reaction, the first pass of the 
algorithm maps the {\tt Theme} to {\tt sbml:reactant} and no {\tt sbml:product} is added. 
For example, E1 (in Example \ref{e:example-1}) would have Akt1 as a {\tt sbml:reactant}. 
To complete this reaction a new {\tt sbml:species} with name phoAkt1 is created representing the 
phosphorylated form of Akt. phoAkt1 is added as the {\tt sbml:product} to the reaction
E1 (See Figure \ref{f:sbml}).

\item[Remove reactions without reactants, products] In other defunct cases
the standoff file might include events that cannot be translated into reactions
with reactants and/or products. For example, we encountered in real data that
a reaction might only have a modifier ({\tt Cause}). Such reactions are automatically 
removed if requested by the user.

\end{description}

\subsection{Implementation}
We used python and the python version of libSBML to 
develop the conversion algorithm. libSBML was used for generating
and accessing the SBML model content. We used a custom 
implementation of a Standoff parser which translates the
line-wise description of standoff triggers and events in a1/a2 and
ann files into a memory structure of triggers (id, type, text) and events
(id, type, roles). These structures are the basis for generating and 
completing the SBML model. The conversion is fast. It scales linearly 
with the number of entities, events and roles.

\subsection{Discussion}
The conversion of standoff format files to SBML is quite straightforward
with a few exceptions where events cannot be mapped directly 
to an SBML reaction as is the case with Localization events that have an 
{\tt AtLoc} role. Moreover, not all entities end up as {\tt sbml:species}. 
Cellular components used in Localization and Transport events, for instance, 
end up as compartments. Another example are Regulation events that have
events as {\tt Theme}. In all of these cases, events in the standoff do not 
have a direct correspondent in the sbml model. 

The algorithm is open to extension. For instance, in order to integrate 
a new event with {\tt Theme}, {\tt Cause}, {\tt Product}, {\tt Site} roles only a new SBO mapping
needs to be defined. 

SBML has graphical editing tool support through, for example, CellDesigner. 
Although CellDesigner uses SBML as its base format, there are a lot
of tool specific custom XML annotations that convey a more fine grained 
view of {\tt sbml:species} and {\tt sbml:reactions} for visualization purposes. 
Our focus in this paper is the conversion to pure SBML format without tool-based 
customizations.

\section{From Event Representations to BioPAX}
Biological Pathway Exchange (BioPAX) is another widely 
used pathway data format based on RDF/OWL. It is used for storage, analysis, 
integration and exchange of pathway models \cite{biopax} . BioPAX, unlike SBML 
is more fine grained in its explicit handling of different types of biological 
players ({\tt bp:PhysicalEntity}\footnote{We will henceforth use the the prefix bp to refer to
BioPAX vocabulary}), and their interactions ({\tt bp:Interaction}).

\begin{figure}
\begin{centering}
\includegraphics[width=.8\columnwidth]{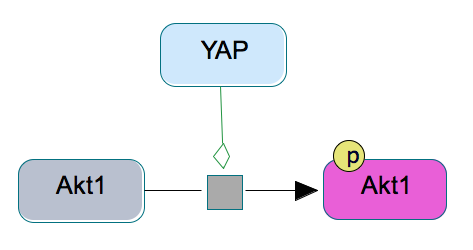}
\caption{Example \ref{e:example-1} converted into BioPAX (viewed with the ChiBE editor)}
\label{f:biopax}
\end{centering}
\end{figure}

The BioPAX conversion algorithm is similar in structure to the SBML conversion. 
Figure \ref{f:biopax} shows Example \ref{e:example-1} converted into a BioPAX model
using the mapping algorithm described in the following paragraphs.

\paragraph{Step 1: Initialize the model}
Read the event files, parse and create a memory internal representation 
of triggers and events. Create an empty model BioPAX model.

\paragraph{Step 2: Create {\tt\bf bp:PhysicalEntity}} 
Each entity in the standoff format is mapped to the corresponding {\tt bp:PhysicalEntity} 
class in BioPAX. {\tt bp:PhysicalEntity} is a superclass of molecules such as 
proteins, DNA, RNA, Small molecules, Complex etc. Depending on the granularity of the 
description of the entity, the element is initialized as a sub class of {\tt bp:PhysicalEntity}. 
The mapping is described in Table 3. The created {\tt bp:PhysicalEntity} 
is assigned a unique-id which is the same as the annotation-ID of the entity. The name 
of the {\tt bp:PhysicalEntity} is assigned the annotation-text. For instance, the protein T8, 
described in the previous section will be encoded as:
 
\begin{small}
\begin{verbatim}
<bp:Protein rdf:about="T8">
 <bp:name rdf:datatype = 
"http://www.w3.org/2001/XMLSchema#string">
IkappaBs</bp:name>
</bp:Protein>
\end{verbatim}
\end{small}

\begin{table}[h]
\begin{footnotesize}
\begin{tabular}{ p{2.8cm}  p{4cm}  }
{\bf Standoff Entity} & {\bf BioPAX class}\\
\hline
Cellular\_component & prefix.CellularLocationVocabulary\\
Complex & prefix.Complex\\
DNA & prefix.Dna\\
Drug & prefix.PhysicalEntity\\
Entity & prefix.PhysicalEntity\\
Gene\_or\_gene\_product & prefix.PhysicalEntity\\
Gene\_product & prefix.PhysicalEntity\\
Gene & prefix.Gene\\
Ion & prefix.PhysicalEntity\\
Protein & prefix.Protein\\
Receptor & prefix.PhysicalEntity\\
RNA & prefix.Rna\\
Simple\_molecule & prefix.SmallMolecule\\
Simple\_chemical & prefix.SmallMolecule\\
Tag & prefix.PhysicalEntity\\
\hline
\end{tabular}
prefix = org.biopax.paxtools.model.level3
\caption{Mapping of Annotation-type to BioPAX term.}
\end{footnotesize}
\label{t:biopax_entity}
\end{table}

\paragraph{Step 3: Create {\tt\bf bp:Interactions}} Each event is mapped 
to the corresponding {\tt bp:Interaction} class in BioPAX. {\tt bp:Interaction} is a 
superclass used to describe reactions and the relationship between the {\tt bp:PhysicalEntity}
elements. Depending on the type of the event, an appropriate sub class of the {\tt bp:Interaction} 
is chosen. The mapping is described in Table \ref{t:biopax_interaction}. The created 
{\tt bp:Interaction} is assigned a unique id which is the same as the annotation-ID of the event
in the standoff. Additionally, all interaction which have the {\tt bp:ConversionDirection} attribute, 
are set to  {\tt bp:LEFT\_TO\_RIGHT}. 

\begin{table}[h]
\begin{footnotesize}
\begin{tabular}{ p{2.8cm}  p{4cm} }
{\bf Standoff Event} & {\bf BioPAX Class} \\
\hline
Conversion  &  prefix.Conversion \\
\hspace{0.2cm}Acetylation  &  prefix.BiochemicalReaction \\
\hspace{0.2cm}Deacetylation  &  prefix.BiochemicalReaction \\
\hspace{0.2cm}Methylation    &  prefix.BiochemicalReaction \\
\hspace{0.2cm}Demethylation    &  prefix.BiochemicalReaction \\
\hspace{0.2cm}Phosphorylation  &  prefix.BiochemicalReaction \\
\hspace{0.2cm}Dephosphorylation  &  prefix.BiochemicalReaction \\
\hspace{0.2cm}Ubiquitination  &  prefix.BiochemicalReaction \\
\hspace{0.2cm}Deubiquitination  &  prefix.BiochemicalReaction \\
Gene\_expression  &  prefix.TemplateReaction \\
\hspace{0.1cm}Transcription  &  prefix.TemplateReaction \\
\hspace{0.1cm}Translation  &  prefix.TemplateReaction \\
Catalysis  &  prefix.Catalysis \\
Degradation  &  prefix.Degradation \\
Catabolism  &  prefix.Degradation \\
Protein\_catabolism  &  prefix.Degradation \\
Association  &  prefix.ComplexAssembly \\
Binding  &  prefix.ComplexAssembly \\
Dissociation  &  prefix.ComplexAssembly \\
Regulation  &  prefix.Control \\
\hspace{0.2cm}Positive\_regulation  &  prefix.Catalysis \\
\hspace{0.3cm}Activation  &  prefix.Control \\
\hspace{0.2cm}Negative\_regulation  &  prefix.Control \\
\hspace{0.3cm}Inactivation  &  prefix.Control \\
Localization  &  prefix.Transport \\
  Transport  &  prefix.Transport \\
\hline
\end{tabular}
prefix = org.biopax.paxtools.model.level3
\caption{Mapping of annotation-type to BioPAX interaction class.}
\end{footnotesize}
\label{t:biopax_interaction}
\end{table}

\paragraph{Step 4: Add participants} 

Events relevant for this paper fall into 3 categories
1) {\tt bp:TemplateReaction} (for transcription ,translation and Gene\_expression events),  
2) {\tt bp:Conversion} (for conversion events including phosphorylation, dephosphorylation etc., 
transport events, binding events and dissociation events) and 3) {\tt bp:Control}  
(for regulation, positive regulation, activation, negative regulation and inactivation events). 

Gene Expression, Transcription and Translation events are modeled as a {\tt bp:TemplateReaction}.
If the {\tt Theme} of a transcription event is of type RNA, then it is mapped to the {\tt bp:product} property of the 
{\tt bp:TemplateReaction}. If the {\tt Theme} is a DNA, then it is added as {\tt bp:template} property.
Similarly, if the {\tt Theme} of a Gene expression event (Translation or Transcription) is of type Protein, 
then the corresponding {\tt bp:PhysicalEntity} is set as the {\tt bp:product} of the {\tt bp:TemplateReaction}. 
If the {\tt Theme} of a Translation event is an RNA, then it is set as the {\tt bp:template} property.

Conversion events are easily mapped to BioPAX elements. Conversion events are all modeled as 
{\tt bp:BiochemicalReaction}.  The {\tt bp:PhysicalEntity} corresponding to {\tt Theme} is set to the 
{\tt bp:left} of the {\tt bp:BiochemicalReaction}. {\tt Site} information is encoded into the suitable 
{\tt bp:sequenceSite} property. 

For instance, in the case of a Phosphorylation event, the reaction corresponds to {\tt Theme} becoming 
phosphorylated. For this a new {\tt bp:PhysicalEntity} is created which has the same properties as 
{\tt Theme}, except that it has an additional {\tt bp:ModificationFeature}, which corresponds to the 
phosphorylated residue. This new entity is then set to {\tt bp:right} of the {\tt bp:BiochemicalReaction}.
If these reactions have the {\tt Cause} entity, then, a new {\tt bp:Control} interaction is created with 
the {\tt Cause} entity as the {\tt bp:controller} and the created {\tt bp:BiochemicalReaction} as the 
{\tt bp:controlled}.

Similarly, Binding, Dissociation and Degradation events map from their definitions onto the BioPAX setting.

Localization and transport events with the {\tt ToLoc} and {\tt FromLoc} roles are handled differently. 
The {\tt ToLoc} and {\tt FromLoc} entities are added as compartments in the BioPAX model. We then 
model a {\tt bp:Transport} reaction with the {\tt Theme} entity transported from the {\tt FromLoc} compartment to the 
{\tt ToLoc} compartment.  Localization events with {\tt AtLoc} role are not explicitly modeled as reaction. Only the compartment 
of the corresponding {\tt Theme}'s {\tt bp:PhysicalEntity} in the BioPAX model is appropriately set. Additionally, the annotation-ID 
of the event is appended as a comment to the corresponding element in BioPAX.

Control events are more complex since they can involve another event as a {\tt Theme} or {\tt Cause}.
Positive/Negative Regulation, Activation and Inactivation events where {\tt Theme} is mapped to
a {\tt bp:PhysicalEntity} are modeled as a {\tt bp:BiochemicalReaction}. Here the entity is converted 
from an active/inactive form to an inactive/active form. Next, a corresponding {\tt bp:Control} interaction 
is created (see Table \ref{t:biopax_interaction}). If the {\tt Cause} is also an entity then it is added 
as the {\tt bp:controller} to the {\tt bp:Control} interaction. However, in case {\tt Cause} is an event, then 
the right side entity (or product) of the Interaction encoded by the {\tt Cause} event is derived and 
added as the {\tt bp:controller}. The previously created {\tt bp:BiochemicalReaction} is then added as the 
{\tt bp:controlled} element for the {\tt bp:Control} interaction. The {\tt bp:controlType} property is set to 
{\tt bp:ACTIVATION} and  {\tt bp:INHIBITION} for the Positive Regulation/Activation and Negative 
Regulation/Inactivation events respectively.

Regulation, Positive Regulation and Negative Regulation can also have events in the {\tt Theme} role. In 
this case, the {\tt Interaction} corresponding to the {\tt Theme} 
is searched, and added as the {\tt bp:controlled} element of a new {\tt bp:Control} interaction. Should there
be a {\tt Cause} entity or event then it is handled as described previously.

\paragraph{Step 5: Optional Postprocessing Operations} 
The software for BioPAX supports post processing similar to the SBML converter:
1) Unused entities can be removed, 2) interactions completed and 3) interactions without
reactants and products removed. Additionally, we can assign a unique identifier to BioPAX 
entities by querying external databases like UniProt, this information is encoded into the 
{\tt bp:Xref} class using either {\tt bp:RelationshipXref} or {\tt bp:UnificationXref}.

\subsection{Discussion}
The conversion from standoff to BioPAX is relatively straightforward. The finer grained 
options to represent different types of information makes it more naturally suited to translate annotations 
from standoff format. Nevertheless, issues highlighted in the SBML conversion exist in the BioPAX 
conversion too. For example, certain events such as Localization events with an {\tt AtLoc} role do
not end up as {\tt bp:Interaction} etc. 

\subsection{Implementation}
The algorithm is implemented in python. It uses the Java Paxtools 4.2.1 toolkit \cite{biopax} to encode and 
manipulate models into the BioPAX format. JPype is used as the bridge to connect python to the Paxtools 
library. The other components of the implementation (such as the standoff-parser) are the same
as used in the SBML implementation. 

\section{Results and Evaluation}

For initial evaluation of our software we used  the mTOR pathway event corpus 
also used in a related study on converting pathway models to standoff 
format \cite{Ohta2011}. The corpus consists of 60 PubMed abstracts and 
the same number of files of hand-annotated standoff files. The 60 abstracts contain 
11960 words. The hand-annotated data contains 1284 events, 1483 Protein, 
1 Entity, 201 Complexes (which gives a total of 2970 text bound annotation triggers). 
In total the annotations contain 1228 {\tt Theme} roles, 19 {\tt Product} roles, 
205 {\tt Causes}, 139 {\tt Site}, 8 {\tt atLoc}, 4 {\tt fromLoc}, 16 {\tt toLoc} and 
51 {\tt participant}. The conversion run on the hand-annotated data correctly 
translates entities and events to SBML and BioPAX according to the mapping 
described in the previous sections.

In order to check our software with state-of-the-art event extraction 
systems we applied an unaltered, freshly downloaded Turku Event Extraction System/TEES Version 2.1
\cite{bjorne2013ddi} to the 60 PubMed abstracts. The resulting TEES/60 
corpus contains 1472 text bound triggers (in a1) and 783 text bound triggers (in a2). TEES 
extracted 1473 Proteins which were all successfully translated to SBML and BioPAX. 20 entities 
were detected, 3 of which were translated into compartments (based on their usage in Localization), 
10 were used as site and translated into site comments. In total 1126 events were detected by TEES of 
which the majority was translated. The exception were 30 localization events of which 1 was a localization 
with an {\tt AtLoc} role (translated into a compartment). 29 Localization events were only annotated with 
a theme and therefore were ignored. 270 regulation events have an event based theme. Only 99 of those are 
also cause annotated and handled as {sbml:reaction}. The remaining 171 disappear since 
the extracted information from TEES is not enough to establish links in the models (both BioPAX and SBML).

Importantly, the failure to translate some of these events into SBML/BioPAX is caused by incomplete information
provided by the NLP event extraction system. For instance, Localization events which only have a {\tt Theme} role
do not provide enough information to be added to the model. Obviously this is one of the areas where hand-annotated 
data provides better conversion results. Nevertheless, these kind of results are encouraging because
the translation into biological knowledge allows for further processing and cleaning of automatically extracted
data and potentially may lead to better extraction systems by providing additional learning signals.

Working with Natural language is never easy. Natural language is full of underspecification,
ambiguities and context-dependencies. Standoff formats represent a compromise between
exact specifications such as SBML and BioPAX that come with their own design approach
and assumptions. Trying to map from one world into the other we noticed a few problems

\begin{description}
\item[Coarse type granularity of biological players:] Coarse granularities such as "Gene or gene product", 
which encompass genes, RNA and protein, make it difficult to assign a type for the entity.
This is important for reactions such as Gene expression, where the decision whether something
is a {\tt sbml:product} or {\tt sbml:modifier} depends on exact distinctions.

\item[Underspecification of event types:]  The event type Regulation 
refers to any process ({\tt Cause}) that modulates any attribute of another process 
({\tt Theme}). In the pathway representation context, it is more natural that the process 
that gets modulated be an event (which is modeled as a {\tt sbml:reaction} in SBML and 
{\tt bp:Interaction} in BioPAX). It is not clear how to correctly represent the scenario when 
the process that gets modulated is an entity (modeled as {\tt sbml:species} in SBML and 
{\tt bp:PhysicalEntity} in BioPAX). However, the event specification allows Theme (that 
which is regulated) to be either an entity or an event.

\item[Underspecification of roles:] 
Event extraction systems try to extract as much as possible
but often are not able to extract all necessary information. For example, the following says
there is a Positive\_regulation on Theme T23, but no information is available on the process that 
is regulating it (no {\tt Cause}).
{\tt E13 Positive\_regulation:T36  Theme:T23}\\
In such cases the converter is unable to extract SBML and BioPAX information. 
\end{description}

\section{Conclusion}

In this paper we proposed and discussed a scheme to convert NLP event representations to standard biomedical 
pathway data formats (SBML and BioPAX). This is important for several reasons. The system allows curators to 
integrate event extraction data into their normal work flow. For instance, the extracted information can give curators 
a base template, which can be further edited in their favorite drawing tool. The  integration into graphical annotation tools 
could provide the basis to later capture the curator's changes. These changes could in turn be used to generate new 
human annotations and to improve current event extraction systems.  Together with other tools that support the conversion of 
SBML models into NLP standoff representations \cite{Ohta2011}, our system bridges the gap between biological modeling 
and automatic event extraction and opens the way to a more tight interaction between the two fields. 

Tight integration of NLP and biomedical research is a recent trend  \cite{huang2015community} with a number of groups moving 
in this direction \cite[ for example]{wei2013pubtator,cejuela2014tagtog,miwa2013method}. For pathway curation, it is important that 
the results of event extraction technologies become part of curation applications/workflows. To achieve this we will have to 
overcome problems inherent in the design of formats such as SBML/BioPAX and/or standoff formats. For instance, SBML wa
s primarily developed as process-based transition notation that cannot faithfully capture all known biochemistry. 
Popular software like CellDesigner add a layer of custom XML annotations to resolve this. For our tools to be used in CellDesigner
we have to add such information in the conversion process. Another layer of information can be provided by automatic annotation 
using UniProt. For the future it will be important to integrate other databases and external references.

Lastly, we plan to perform a more thorough evaluation of the conversion by reconstructing a complete known pathway 
(e.g. the mTOR pathway, for which high quality maps are already available). We are also performing a large scale evaluation 
of the software on the EVEX event database -- a text mining resource of PubMed abstracts and 
full texts \cite{van2011evex}

\bibliographystyle{acl}
\bibliography{papers}

\end{document}